\documentclass[letterpaper]{article} 
\usepackage{aaai25}  
\usepackage{times}  
\usepackage{helvet}  
\usepackage{courier}  
\usepackage[hyphens]{url}  
\usepackage{graphicx} 
\usepackage{natbib}  
\usepackage{caption} 
\nocopyright
\usepackage{algorithm}
\usepackage{algorithmic}
\usepackage{booktabs}
\usepackage{tabularx}
\usepackage{array}
\newcolumntype{L}{>{\raggedright\arraybackslash}X}
\newcolumntype{C}{>{\centering\arraybackslash}X}
\usepackage{amsmath,amsthm,amssymb,amsfonts}
\usepackage{makecell}
\usepackage{bm}

\usepackage{newfloat}
\usepackage{listings}
\DeclareCaptionStyle{ruled}{labelfont=normalfont,labelsep=colon,strut=off} 
\floatstyle{ruled}
\newfloat{listing}{tb}{lst}{}
\floatname{listing}{Listing}
\title{Lightweight Multiplane Images Network for Real-Time Stereoscopic Conversion from Planar Video}
\author{
    Shanding Diao\textsuperscript{\rm 1},
    Yang Zhao\textsuperscript{\rm 1,2},
    Yuan Chen\textsuperscript{\rm 3},
    Zhao Zhang\textsuperscript{\rm 1},
    Wei jia\textsuperscript{\rm 1},
    Ronggang Wang\textsuperscript{\rm 2,4}
}
\affiliations{
    \textsuperscript{\rm 1}School of Computer and Information, Hefei University of Technology, Hefei 230009, China\\
    \textsuperscript{\rm 2}Peng Cheng National Laboratory, Shenzhen 518000, China\\
    \textsuperscript{\rm 3}School of Internet, Anhui University, Hefei 230039, China\\
    \textsuperscript{\rm 4}School of Electronic and Computer Engineering, Peking University Shenzhen Graduate School,  Shenzhen 518055, China\\
    yzhao@hfut.edu.cn, 
    rgwang@pkusz.edu.cn
}
\usepackage{bibentry}

\begin{document}

\maketitle

\begin{abstract}
With the rapid development of stereoscopic display technologies, especially glasses-free 3D screens, and virtual reality devices, stereoscopic conversion has become an important task to address the lack of high-quality stereoscopic image and video resources. Current stereoscopic conversion algorithms typically struggle to balance reconstruction performance and inference efficiency. This paper proposes a planar video real-time stereoscopic conversion network based on multi-plane images (MPI), which consists of a detail branch for generating MPI and a depth-semantic branch for perceiving depth information. Unlike models that depend on explicit depth map inputs, the proposed method employs a lightweight depth-semantic branch to extract depth-aware features implicitly. To optimize the lightweight branch, a heavy training but light inference strategy is adopted, which involves designing a coarse-to-fine auxiliary branch that is only used during the training stage. In addition, the proposed method simplifies the MPI rendering process for stereoscopic conversion scenarios to further accelerate the inference. Experimental results demonstrate that the proposed method can achieve comparable performance to some state-of-the-art (SOTA) models and support real-time inference at 2K resolution. Compared to the SOTA TMPI algorithm, the proposed method obtains similar subjective quality while achieving over $40\times$ inference acceleration.
\end{abstract}

\begin{figure*}
  \includegraphics[width=\linewidth]{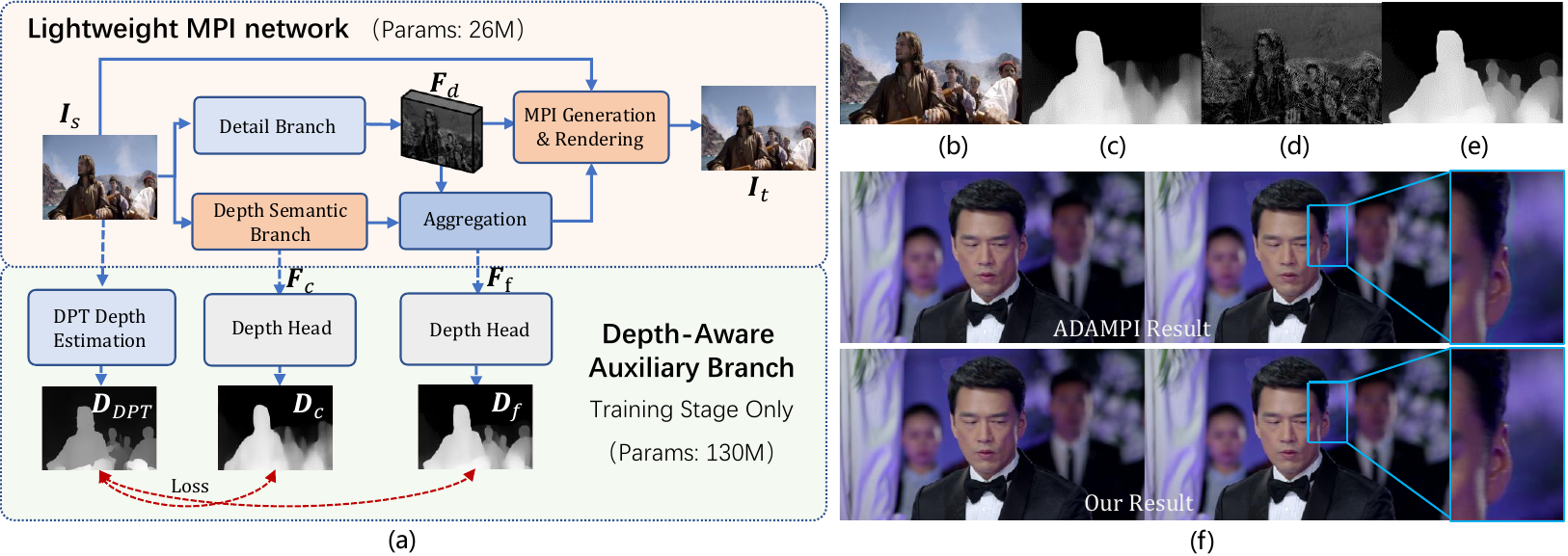}
  \caption{The proposed lightweight multiplane images stereoscopic conversion network, (a) overall framework, (b) input image, (c) coarse depth map predicted by depth head, (d) visualization example of detail features, (e) fine-grained depth map, (f) 3D video conversion results of ADAMPI and the proposed method.}
  \vspace{-5pt}
  \label{fig:1}
\end{figure*}

\section{Introduction}
With the development of ultra-high-definition (4K/8K) and high-dynamic-range display devices, planar video display has approached the limits of human visual perception. To further enhance the visual experience, immersive displays represented by stereoscopic video (3D), virtual reality (VR), and free-viewpoint video are needed. In the past years, with the rapid progress of autostereoscopic displays, commonly known as glasses-free 3D screens \cite{hua2022recent}, stereoscopic displays have received considerable attention from both industry and academia. However, the scarcity of stereoscopic video resources has become one of the bottlenecks restricting the development of the stereoscopic display industry. Therefore, high-quality algorithms for converting planar video to stereoscopic video (3D video conversion) have become an important research direction.

For planar-to-stereo conversion, an extra viewpoint is created that, in conjunction with the original viewpoint, mimics the distinct images captured by two eyes \cite{steffen2019neuromorphic, read2022stereo}. Most stereoscopic conversion methods are predicated on the utilization of disparity warping. For instance, the traditional depth-image-based rendering (DIBR) \cite{fehn2004depth} employs a depth map from the original viewpoint to render another viewpoint. However, the performance of DIBR-based algorithms highly depends on the accuracy of depth maps, which are usually obtained through manual creation or monocular depth estimation methods \cite{ranftl2021vision, wofk2019fastdepth}, often leading to depth errors and occlusion-exposed hole artifacts. Some subsequent methods adopt deep neural networks to improve predicted view \cite{xie2016deep3d, lee2017automatic}. However, these approaches struggle to reconstruct the 3D structure and dense geometry of various scenes accurately.

Compared to traditional planar-to-stereo conversion networks, multiplane images (MPI)-based methods do not explicitly utilize depth/disparity to warp pixels. Instead, they map the 3D spatial scene into several fronto-parallel planes and then synthesize novel view through MPI rendering, which offers robustness against errors in estimated depth maps and naturally avoids hole-filling problems. Although inferring an MPI representation from a monocular image remains challenging, planar images or videos are the most prevalent and common contents, leading to ongoing research in single-view MPI methods.
For example, MINE \cite{li2021mine} utilized the predicted MPI representation to render depth maps and then calculated the loss against ground truth (GT) depth maps to implicitly infer depth information. However, the lack of direct depth cues may limit its effectiveness. Subsequent single-image MPI methods \cite{han2022single} utilize both scene images and corresponding depth maps to calculate the MPI representation. Recently, temporal multiplane images (TMPI) \cite{diao2024stereo} extended MPI representation by incorporating temporal information from adjacent frames to recover the missing details in occlusion-exposed regions, and thus obtain the state-of-the-art (SOTA) performance in 3D video conversion. However, introducing temporal information further increases the computational cost of TMPI. Although MPI-based 3D video conversion algorithms can achieve high-quality visual results, the high computational cost significantly hampers their practical applications.

To achieve real-time 3D video conversion, this paper proposes a lightweight MPI stereoscopic conversion network (LMPIN), which mainly consists of a detail feature branch, a depth semantic feature branch, and a light MPI rendering module. The proposed method adopts a heavy training and lightweight inference strategy, where an additional depth-aware auxiliary branch is introduced during the training phase to assist in learning depth information. As shown in Fig.\ref{fig:1}(a), since the proposed model does not explicitly perform monocular depth estimation, the auxiliary branch employs a depth head to produce a coarse depth map (Fig.\ref{fig:1}(c)) from depth semantic features, and then use a second depth head to estimate the refined depth map (Fig.\ref{fig:1}(e)) by fusing detail features. Furthermore, a large-scale pretrained monocular depth estimation model is used to obtain a reference depth map to constrain the depth maps in the coarse-to-fine refinement process. Finally, improvements are further made to accelerate the MPI rendering process. Fig.\ref{fig:1}(f) illustrate the planar-to-stereo conversion results of ADAMPI \cite{han2022single} and our method, respectively, which show the proposed method can obtain better subjective quality with a much lighter structure.

The main contributions can be summarized as follows:
\begin{itemize}
\item This paper proposes a lightweight architecture to predict MPI for 3D video conversion. Compared to conventional single-image MPI models that require an additional monocular depth estimation network to provide depth maps, the proposed method only adds a lightweight depth semantic branch to implicitly perceive the depth of the scene and greatly reduces computational overhead.
\item A depth-aware training auxiliary branch is introduced to learn the perception of depth information, which adopts a coarse-to-fine structure and uses a pretrained largescale depth estimation model to obtain supervision depth maps. This auxiliary branch is only calculated during training and thus accelerates the inference process.
\item This paper introduces a light rendering approach specifically suited for 3D conversion that efficiently generates high-resolution images. Experimental results demonstrate that the proposed method can achieve high-quality and real-time 3D video conversion for 2K resolution. 
\end{itemize}

\section{Related Work}
\subsection{Planar-to-Stereo Conversion}
Typically, the process of converting 2D contents to stereo consists of two interconnected steps \cite{fehn2004depth, xie2016deep3d, Zhang_Wang_2022}. 
The first step is to define the depth structure of the scene, which usually requires the creation of a depth map.
In the second step, the estimated depth information and original texture content are used to generate a novel view through different rendering techniques, thereby forming a stereoscopic image pair.
When considering other view synthesis tasks, it becomes evident that several methodologies, such as layered depth image (LDI) \cite{tulsiani2018layer, shih20203d} and MPI \cite{tucker2020single, li2021mine}, can be employed to characterize a 3D scene instead of merely a depth map. These approaches typically lead to a more comprehensive understanding of the scene structure, thereby yielding better stereo results. In these methods, it is usually necessary to input a depth map as the source of depth information when there is only a single viewpoint is available. This depth map is generally obtained through manual labeling or monocular algorithms, resulting in additional computation. Hence, Some lightweight monocular depth estimation algorithms attempt to reduce the cost of depth estimation by employing lightweight backbone \cite{wofk2019fastdepth}, network pruning \cite{cheng2023survey}, and knowledge distillation \cite{song2023knowledge}.

\begin{figure*}
  \includegraphics[width=\linewidth]{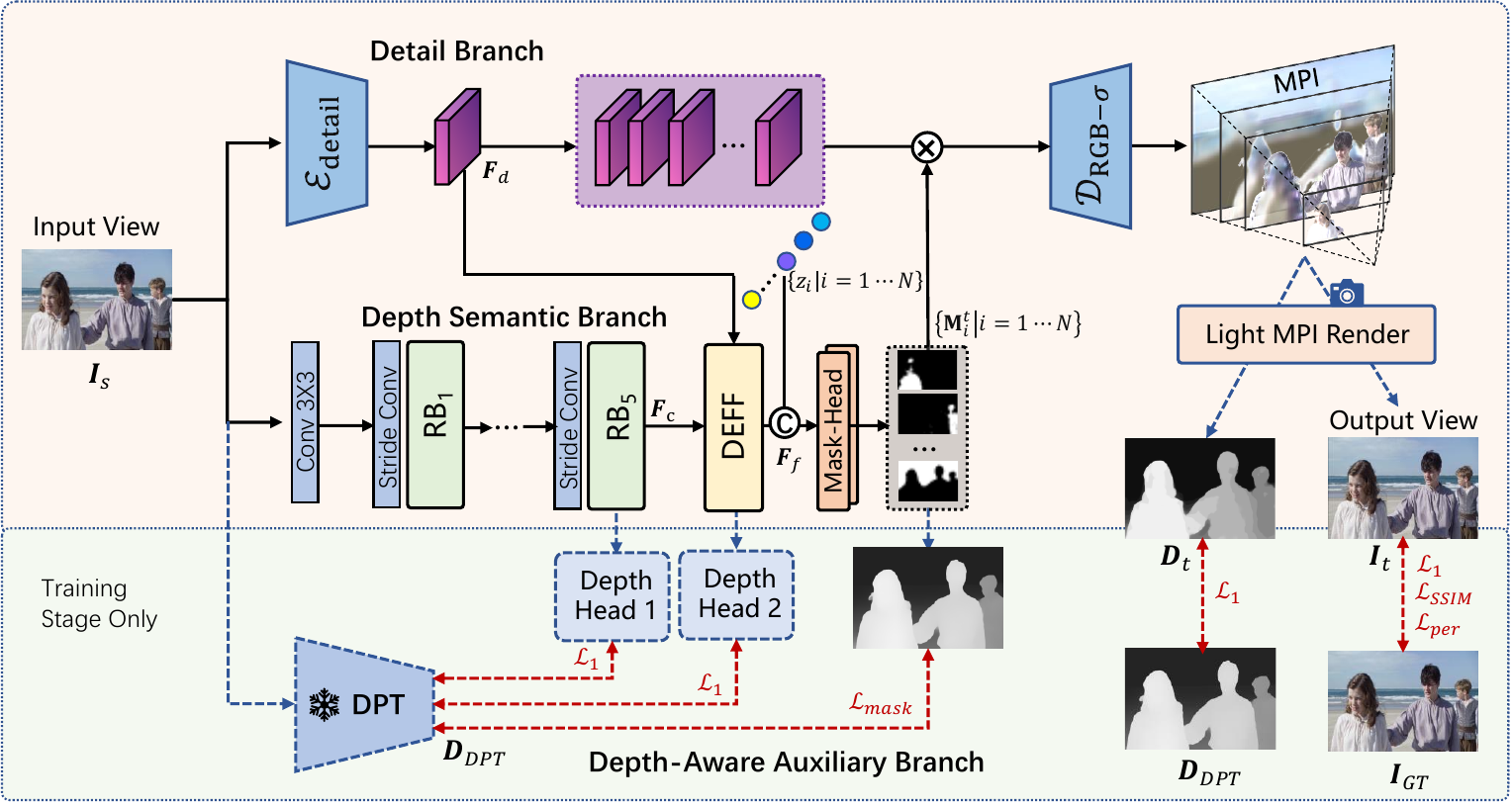}
  \caption{Architecture of the proposed lightweight multiplane images stereoscopic conversion network.}
  \vspace{-5pt}
  \label{fig:2}
\end{figure*}

\begin{figure}
  \includegraphics[width=\linewidth]{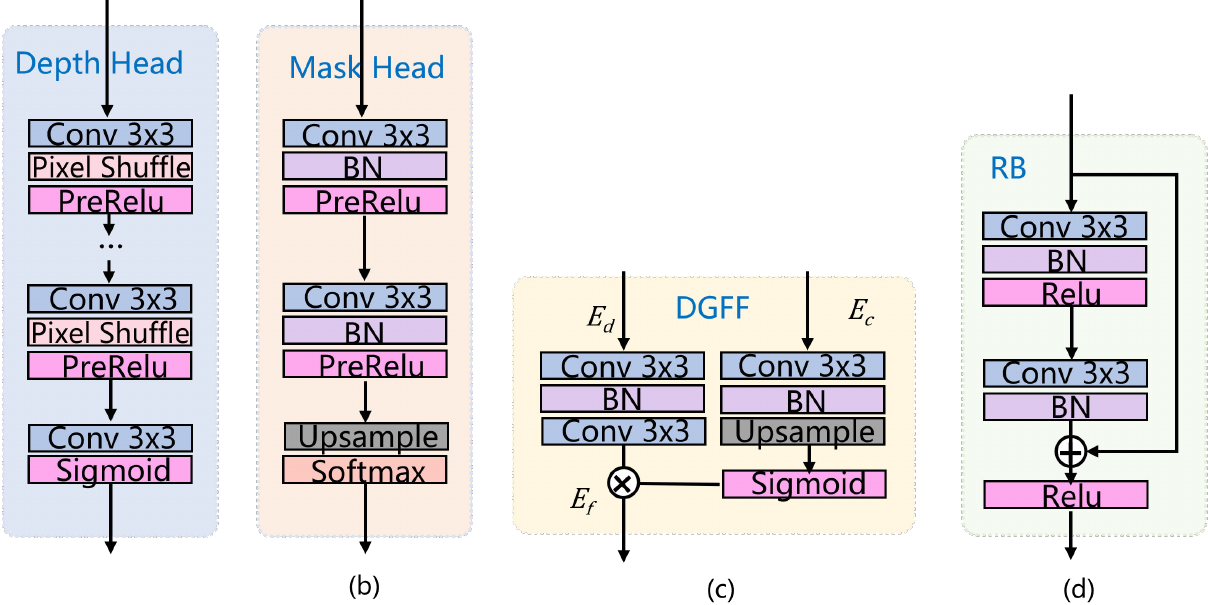}
  \caption{The structure of different blocks. (a) depth head, (b) mask head, (c) depth-guided enhanced feature fusion (DEFF) block, (d) basic residual blocks (RB)}
  \vspace{-15pt}
  \label{fig:3}
\end{figure}

\subsection{3D representation model}
In recent years, numerous SOTA 3D representation algorithms have been proposed. Zhou et al. \cite{zhou2018stereo} introduced MPI, representing scenes as fronto-parallel planes at fixed depths. While MPI is highly generalizable, it remains a 2.5D representation, susceptible to imperfections when observed from non-frontal views. Neural Radiance Fields (NeRF) \cite{mildenhall2021nerf} employ fully connected deep networks to implicitly model scenes. NeRF captures fine details from any view but demands substantial computational resources and lacks generalization to new scenes. The Large Reconstruction Model (LRM) \cite{hong2023lrm}, based on transformers and trained on millions of 3D models, directly predicts NeRF representations from single images. LRM excels in generalization but struggles to handle complex shapes and backgrounds. Recently, 3D Gaussian Splatting (3DGS) \cite{kerbl20233d} has received lots of attention, which represents scenes with Gaussian spheres and can efficiently synthesize novel views through splatting rendering. However, 3DGS still needs to optimize high-quality 3D representation for each scene thus leading to high cost and low generalization for large scene and high-resolution images. 
In our exploration of 3D representation for planar-to-stereo conversion, MPI shows greater potential for real-time stereoscopic conversion. Other models, which are more suitable for 3D reconstruction and modeling from arbitrary viewpoints, are considerably more time-consuming. For stereoscopic videos observed only from frontal views, using a simplified MPI to represent the spatial scene is more concise and better suited for the task.

\section{Method}
For each single-view frame $\bm{I}_s \in \mathcal{R}^{H\times W \times 3}$, our objective is to synthesize an MPI representation $\left\{\bm{P}_s^n\in{\mathcal{R}}^{H\ \times\ W\ \ \times4}|\ n\ =\ 1\ \cdots\ N\right\}$ to characterize the spatial information, where $N$ denotes the total number of planes. Subsequently, a novel target novel view $\bm I_t$ can be rendered from the MPI representation. 

\subsection{Multi-Plane Images for 3D Conversion}
To generate an MPI representation for frame $\bm I_s$, we propose a lightweight MPI network (LMPIN) to generate $N$ fronto-parallel RGB-$\sigma$ planes. Each plane $\bm P_s^n$ consists of three-channel color map $\bm{C}_s^n$ and one-channel density map $\bm{\sigma}_s^n$ which are derived from ${\bm I}_s$ and corresponding plane depth $z_n$\ as:
\begin{equation}
\label{eqn:01}
(\bm{C}_s^n,\bm{\sigma}_s^n)\ =\ LMPIN({\ \bm{I}}_s,\ z_n).   
\end{equation}

We then employ pixel warping from the source MPI representation in a differentiable manner. Because 3D video conversion mainly focuses on horizontal disparity, each pixel $(x_t,y_t)$ at $n$-th target MPI plane $\bm P_t^n$ can be mapped to pixel $(x_s,y_s)$ on $\bm P_s^n$ via simplified homography function:
\begin{equation}
\label{eqn:02}
\left[\begin{matrix} x_s \\ y_s \\ 1 \end{matrix}\right]^T \bm{K} \left( \bm{I} - \frac{\bm{t} \bm{n}^T}{z_{\bm{n}}} \right) \bm{K}^{-1} \left[\begin{matrix} x_t \\ y_t \\ 1 \end{matrix}\right]^T
\end{equation}
where $\bm{t}$ denotes the translation matrix from the source viewpoints to the target viewpoints, $\bm{K}$ is the camera intrinsic, $\bm{I}$ represents an identity matrix, and $n = [0, 0, 1]$ is the normal vector. 
The predicted target view $\bm{I}_t$ is then obtained by alpha compositing the color images in back-to-front order using the standard over operation as in \cite{Porter_Duff_1984}:
\begin{equation}
\label{eqn:03}
\bm{I}_t\ =\ \sum\nolimits_{n=1}^{N}{(\bm{C}_t^n\bm{\alpha}_t^n\prod\nolimits_{j=1}^{n-1}{(1-\bm{\alpha}_t^j}\ ))},    
\end{equation}
where $\bm{\alpha}_t^n=\ exp(-\bm{\delta}_t^n\bm{\sigma}_t^n)$  and $\bm{\delta}_n$ is the distance map between $n$-th and $(n+1)$-th planes, and we set the depth of MPI planes uniformly spaced in disparity.

\subsection{Light MPI Network}
As shown in Fig.\ref{fig:2}, the proposed network consists of a detail branch, a depth semantic branch, an MPI rendering module, and an extra depth-aware training auxiliary branch. In the following, details of each branch will be introduced. The detail branch is responsible for generating the context of each plane for the MPI representation. A commonly used auto-encoder \cite{zhang2023lite} is adopted for the detail branch, which is a multi-scale encoder-decoder structure. Note that the output of the decoder has been adjusted to produce 4-channel maps. The encoder $\mathcal{E}_{detail}$ is utilized to extract spatial detail features ${\bm{F}}_d$ only once per image,
\begin{equation}
\label{eqn:04}
{\bm{F}}_d\ =\mathcal{E}_{detail}({\bm{I}}_s).    
\end{equation}

The depth semantic branch is used to perceive the scene depth information. Since a large receptive field is important for global depth perception, we alternately use convolution with a stride of 2 for downsampling and employ basic residual blocks (RBs) \cite{he2016deep} for feature processing. The depth semantic branch uses 5 downsampling operations, effectively enlarging the receptive field and reducing the computational cost. The final RB outputs preliminary depth-semantic features ${\bm{F}}_c$. Since the ${\bm{F}}_c$ is computed in low-resolution space, the features are coarse and lack accurate edge details. Inspired by classical coarse-to-fine structure in semantic segmentation task \cite{yu2018bisenet, yu2021bisenet}, a depth-guided enhanced feature fusion (DEFF) block is presented, as shown in Fig.\ref{fig:3}(c). This fusion block $f_{DEFF}$ utilizes upsampled coarse depth features as attention to guide the fusion of detail features ${\bm{F}}_d$ from the detail branch, and then produce the fine depth semantic features ${\bm{F}}_f$, as follows,
\begin{equation}
\label{eqn:05}
\bm{F}_f\ =f_{DEFF}({\ \bm{F}}_d,\ {\ \bm{F}}_c) .  
\end{equation}

The fine depth semantic features are further concatenated with plane depth values $z_n$, and then feed into multiple mask heads to generate the assign masks $\left\{\bm{M}_n\in\mathbb{R}^{H\ \times\ W\ \ \times\ 1}|\ n\ =\ 1\ \cdots\ N\right\}$ that segments the image into different planes according to depth values, as:
\begin{equation}
\label{eqn:06}
\bm{M}_n=f_{Mask}\left({\ \bm{F}}_f \oplus{\ \bm{Z}}_n\right),\ \ n\ =\ 1\ \cdots\ N,\    
\end{equation}
where $\oplus$ denotes concat operation, ${\bm{Z}}_n$ represents a depth value map with all values are ${z}_n$, and $f_{Mask}$ denotes the mask head. The structure of the mask head is shown in Fig.\ref{fig:3}(b), which sequentially contains two convolution layers with a bilinear upsampling layer and a Softmax layer.

To constrain the lightweight depth-semantic branch for better perception of image depth information, a depth-aware auxiliary training branch is employed, which also follows a coarse-to-fine structure. As illustrated in Fig.\ref{fig:2}, two depth heads are employed to output coarse depth map ${\bm{D}}_c$ and fine depth map ${\bm{D}}_f$ from coarse features ${\bm{F}}_c$ and fine features ${\bm{F}}_f$, respectively. The structure of the depth head is shown in Fig.\ref{fig:3}(a). Subsequently, a pre-trained large-scale monocular depth estimation model, DPT \cite{ranftl2021vision}, is utilized to obtain a reference depth map ${\bm{D}}_{DPT}$. By constraining the similarity between the depth maps ${\bm{D}}_c$, ${\bm{D}}_f$ and the reference depth ${\bm{D}}_{DPT}$, the depth-aware auxiliary branch can improve the learning of depth information. As mentioned earlier, this auxiliary branch is only used in the training phase and does not introduce additional depth estimation computations during inference. 

After obtaining the assign masks, the encoded detail features ${\bm{F}}_d$  are replicated $N$ times and obtain the features for each plane through multiplication with the assign masks $\bm{M}_n$. The RGB-$\sigma$ decoder $\mathcal{D}_{RGB\sigma}$ finally runs $N$ times, decoding these features into $N$ front-parallel planes$ \left\{\bm{P}_s^n\in\mathbb{R}^{H\ \times\ W\ \ \times4}|\ n\ =\ 1\ \cdots\ N\right\}$, as follows,
\begin{equation}
\label{eqn:07}
\bm{P}_s^n\ =\ \mathcal{D}_{RGB\sigma}\left({\ \bm{F}}_d\ \ast\sum\nolimits_{j=n}^{N}{\bm{M}_j\ \ }\right), 
\end{equation}
where $\sum_{j=n}^{N}{M_j}$ calculates the combination of the pixels on and behind the $n$-th plane, which denotes the context regions for planes $\bm{P}_s^n$. 

\subsection{Accelerate Rendering with Low-Resolution MPI }
 MPI-based methods \cite{tucker2020single, li2021mine, han2022single} typically blend the input image ${\bm{I}}_s$ with the predicted color map $\bm{C}_s^n$ for each plane during rendering. They assume that visible content should use the foreground image ${\bm{I}}_s$, while occluded content should rely on the network-predicted color map. Consequently, the blend weight $\bm{w}_n$ can be calculated through the cumulative multiplication of opacity, as:
\begin{equation}
\label{eqn:09}
\bm{w}_n\ =\prod\nolimits_{j>n}^N{(1\ - \alpha_s^n)},
\end{equation}
A larger value in $\bm{w}_n$ indicates no obstruction in front, and thus, a greater inclination towards using the foreground image, and vice versa.

In real-world stereo video conversion applications, the resolution of video resources usually reaches 2K ($1920\times 1080$) or larger. This presents a significant computational challenge for predicting MPI representation. 
In our approach, we compute the MPI in low-resolution space to accelerate MPI calculation and rendering, each plane consisting of a low-resolution color map and a density map, denoted as $(\bm{C}_\downarrow^n,\bm{\sigma}_\downarrow^n)$. Subsequently, these low-resolution planes are magnified to the same size as the original image ${\bm{I}}_s$ through bilinear upsampling $u_\uparrow(\cdot)$. Then, the final color map can be calculated as, 
\begin{equation}
\label{eqn:10}
{\bm{C}^\prime}_s^n\ =\ u_\uparrow(\bm{w}_n){\ \bm{I}}_s\ +\ (1\ - u_\uparrow(\bm{w}_n))\ {\ u_\uparrow(\bm{C}_\downarrow^n)}. 
\end{equation}

In 3D video conversion, the pixel values of the synthetic viewpoint are warped from the high-resolution planar image. Thus, reducing the resolution of MPI does not lead to resolution distortion in the synthetic view directly. In addition, for the small and smooth occlusion regions, the artifacts introduced by upsampling of MPI are not prominent. Consequently, this strategy allows for efficient calculation while maintaining the quality of the synthesized output.

\subsection{Loss Function}
The loss function of the proposed method consists of two parts. The first part is the depth information loss $\mathcal{L}_{depth}$, used to constrain the learning of the depth-semantic branch. The other part is the MPI loss $\mathcal{L}_{MPI}$, which supervises the learning of MPI by rendering the target viewpoint and comparing the target viewpoint image to the GT image. 

For the computation of $\mathcal{L}_{depth}$, we employ ${\mathcal{L}}_1$ loss in the auxiliary training branch to constrain the similarity between the predicted depth maps and the reference depth map from the DPT \cite{ranftl2021vision}. Inspired by the ADAMPI \cite{han2022single}, the mask loss is also used to constrain the consistency between multi-layer masks and the reference depth map, as follows:
\begin{equation}
\label{eqn:11}
\mathcal{L}_{depth}\ =\ {\mathcal{L}}_1\left({\bm{D}}_c,{ \bm{D}}_{DPT}\right)+\mathcal{L}_1\left({\bm{D}}_f,{\ \bm{D}}_{DPT}\right)+\lambda\mathcal{L}_{mask},  
\end{equation}
\begin{equation}
\label{eqn:12}
\mathcal{L}_{mask}\ =\ \frac{1}{HW}\sum_{n=1}^{N}{\sum_{(x,y)}\bm{M}_n\ast\ |\bm{D}_{DPT}-z_n|},
\end{equation}
where the weight $\lambda$ is experimentally set as 10 so that these three terms are of the same order of magnitude. 
For computing MPI loss, we use commonly used L1 loss, SSIM loss, and perceptual loss \cite{liu2018photographic} to jointly constrain the consistency between the predicted target viewpoint image $\bm{I}_t$ and the GT image $\bm{I}_{GT}$. Additionally, the reference depth map $\bm{D}_{DPT}$ of the target view is also used to constrain the depth map $\bm{D}_t$ of the predicted viewpoint, aiding in the optimization of depth and content, where $\bm{D}_t$ is rendered using the optimized MPI. The MPI loss $\mathcal{L}_{MPI}$ and the final total loss $\mathcal{L}_{total}$ are defined as,
\begin{equation}
\label{eqn:13}
\begin{split}    
\mathcal{L}_{MPI}\ =\ {\ \mathcal{L}}_1\left(\bm{I}_t,\bm{I}_{GT}\right)+{\ \mathcal{L}}_{SSIM}\left(\bm{I}_t,\bm{I}_{GT}\right)+\\ {\ \mathcal{L}}_{per}\left(\bm{I}_t,\bm{I}_{GT}\right)+\ \mathcal{L}_{1}\left(\bm{D}_t,\bm{D}_{DPT}\right), 
\end{split}
\end{equation}
\begin{equation}
\label{eqn:14}
\mathcal{L}_{total}\ =\ {\ \mathcal{L}}_{depth}\ +{\ \mathcal{L}}_{MPI}.
\end{equation}

\begin{figure*}
  \includegraphics[width=\linewidth]{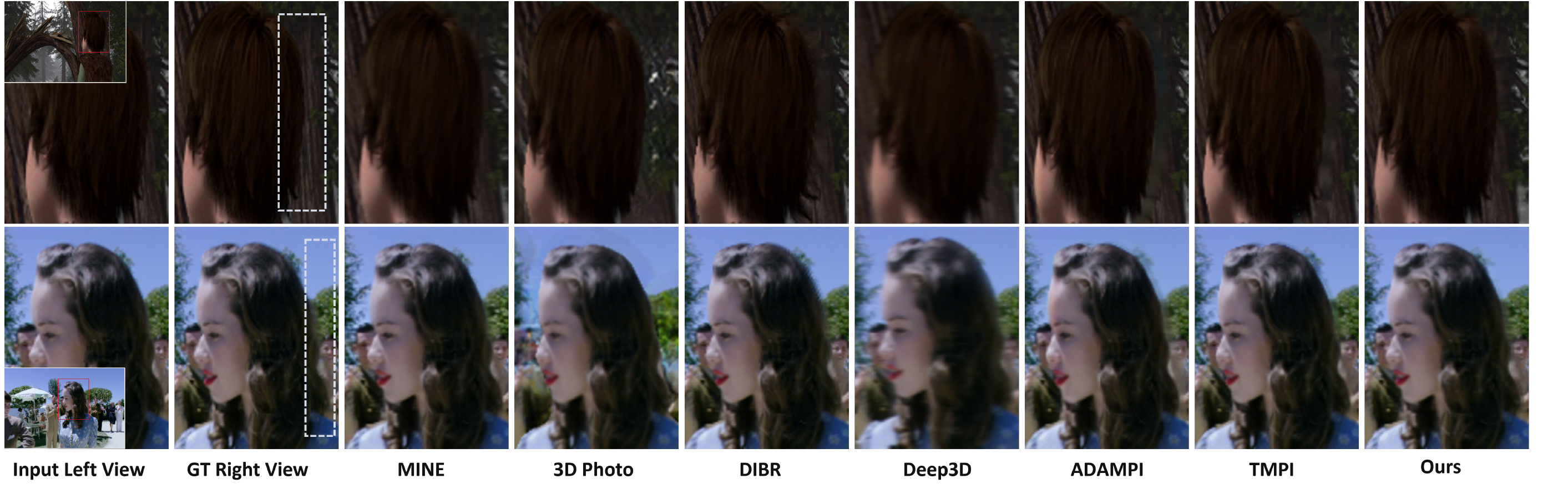}
  \caption{The planar-to-stereo conversion results of different methods on 3D movie test set.}
  \vspace{-5pt}
  \label{fig:4}
\end{figure*}

\section{Experiments}
\subsection{Datasets and Implementation Details}
\subsubsection{Training set}
To reproduce 3D videos with diverse content, the training set must exhibit high diversity by including a wide range of scene types. However, existing stereoscopic image datasets often lack both the quantity of images and the variety of scenes. To address this, we construct a synthetic training set based on the large-scale COCO dataset \cite{caesar2018coco}. Following the data preparation method as in \cite{watson2020learning}, we first predict disparity from a single image and then use the estimated disparity to generate stereo pairs. In our experiments, we utilize a total of 111K pairs of images with rich diversity for training. We have rescaled the resolution of all training pairs to $256\times 384$.

\subsubsection{Test set}
For testing, we use the same authentic 3D movie test set as in TMPI \cite{diao2024stereo}, which contains a total of 3,323 five-frame sequences. We utilized the left view as input and reconstructed the right view employing different methods. In addition, to verify the generalization and robustness of high-resolution 3D video conversion, we randomly select 10 2K ($1920\times1080$) planar videos from the Youku video super-resolution and enhancement dataset (Youku2K) \cite{YoukuVSRE2019}, which are similar to the contents in planar-to-stereo application scenarios. Then, the performance and inference speeds of different methods are tested on this set.

\subsubsection{Implementational details}
Due to the difficulty of simultaneously optimizing depth perception and MPI generation, we initially pretrain the encoder $\mathcal{E}_{detail}$, depth semantic branch, and auxiliary branch for 200,000 steps specifically for preliminary depth information perception. Subsequently, the entire network is jointly trained for an extensive 800,000 steps with an initial learning rate of 0.0002 for the encoder, 0.001 for the decoder $\mathcal{D}_{RGB\sigma}$, and 0.00001 for the depth semantic branch. Note that the learning rate for the decoder is larger than other terms, as the other modules have already undergone initial optimization, and the reconstructed MPI of the decoder ultimately determines the quality of the final output image.  
The model adopts the Adam optimizer with a weight decay of 1e-4 during the training stage. The number $N$ of planes was set to 16 due to the disparity between left and right views is not large in stereoscopic videos.

\subsection{Experimential Results}
To verify the effectiveness of the proposed method, comparisons are conducted with several SOTA MPI-based models of ADAMPI \cite{han2022single}, MINE \cite{li2021mine}, and TMPI \cite{diao2024stereo}, other 3D conversion networks of Deep3D \cite{xie2016deep3d} and 3D-Photo \cite{shih20203d}, and traditional DIBR technique.

\subsubsection{3D Video Conversion Results}

Fig.\ref{fig:4} illustrates the planar-to-stereo conversion results of different methods. Firstly, it is noteworthy that MINE, Deep3D, and the proposed method do not utilize extra depth maps. For those methods that need depth inputs, we use the same pretrained DPT \cite{ranftl2021vision} model to predict depth maps. Secondly, traditional DIBR tends to leave numerous holes and artifacts along object edges, even after inpainting operations. The 3D-Photo may inpaint false contents in occlusion-exposure regions. Thirdly, horizontal translation and fusion strategy in Deep3D leads to blurry details. Lastly, by comparing the MPI-based models, MINE produces smaller disparities than other methods, and the proposed lightweight model can obtain comparable subjective results to SOTA TMPI and efficiently avoid visual artifacts around the foreground.

\begin{figure}[!t]
  \includegraphics[width=\linewidth]{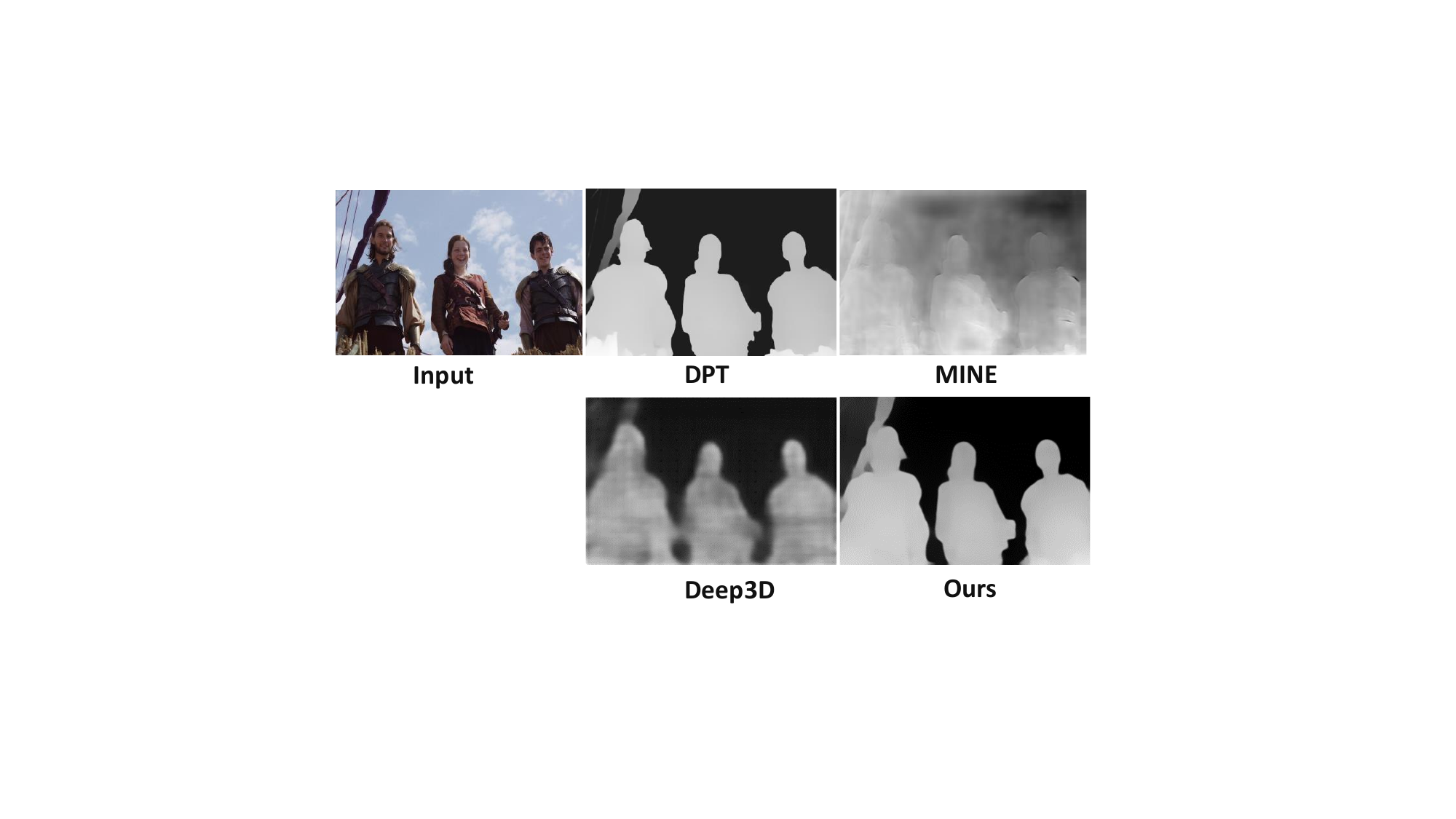}
  \caption{ Comparative results of depth map generation using various methods.}
  \vspace{-10pt}
  \label{fig:5}
\end{figure}

Fig.\ref{fig:5} further shows the depth maps generated by several methods that do not explicitly utilize depth maps. Compared with the depth map estimated by DPT, it can be observed that the depth information learned by the proposed method is already quite accurate. This indicates that the proposed depth-semantic branch and the auxiliary training branch are capable of perceiving reasonable depth and accurate boundary information, which can further promote the generation of high-quality MPI.

For objective testing, three commonly used assessments are used, i.e., basic distortion metrics PSNR and SSIM, and one perceptual similarity measure LPIPS \cite{zhang2018unreasonable}. Table~\ref{table:2} lists the quality scores of these methods. It is observed that the proposed method surpasses MINE and Deep3D which do not employ depth maps as input across all metrics. Remarkably, the proposed method can achieve comparable LPIPS scores to SOTA ADAMPI and TMPI models with much lighter structure and fewer parameters.

\begin{figure}[!t]
  \includegraphics[width=\linewidth]{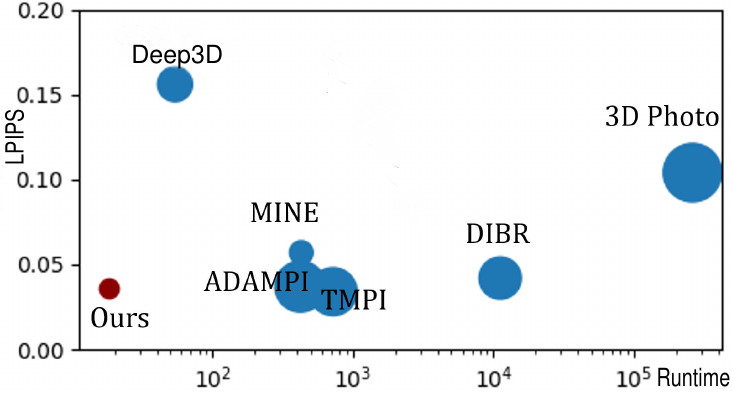}
  \caption{LPIPS perception quality and runtime planes of different methods. The size of each dot represents the size of its parameters.}
  \label{fig:6}
  
\end{figure}

Fig.\ref{fig:6} visually compares the perceptual metric LPIPS, runtime, and parameters of different methods. We can find that the proposed lightweight model can achieve high-quality 3D conversion results in a more efficient way.
\begin{table}[!t]
\small
\centering
\begin{tabular}{lcccccc}
\toprule
Method   & Extra       & SSIM↑                       & PSNR↑           & LPIPS↓         & Param(M)    \\
\midrule 
ADAMPI   & DPT         & 0.923                       & 33.307          & 0.037          & 57+123              \\
MINE     & -           & 0.877                       & 30.780          & 0.057          & 38                    \\
3D Photo & DPT         & 0.902                       & 29.735          & 0.104          & 114+123          \\
Deep3D   & -           & 0.830                       & 28.567          & 0.156          & 84                   \\
DIBR     & DPT         & 0.892                       & 32.929          & 0.042          & 123           \\
TMPI     & DPT         & \textbf{0.924}              & \textbf{33.630} & \textbf{0.034} & 37+123             \\
Ours     & -           & 0.913                       & 33.037          & 0.036          & \textbf{26}  \\
\bottomrule
\end{tabular}
\caption{ PSNR, SSIM, LPIPS scores and Parameters of different methods on the 3D movie test set.}
\vspace{-10pt}
\label{table:2}
\end{table}
\begin{figure*}[!t]
  \includegraphics[width=\linewidth]{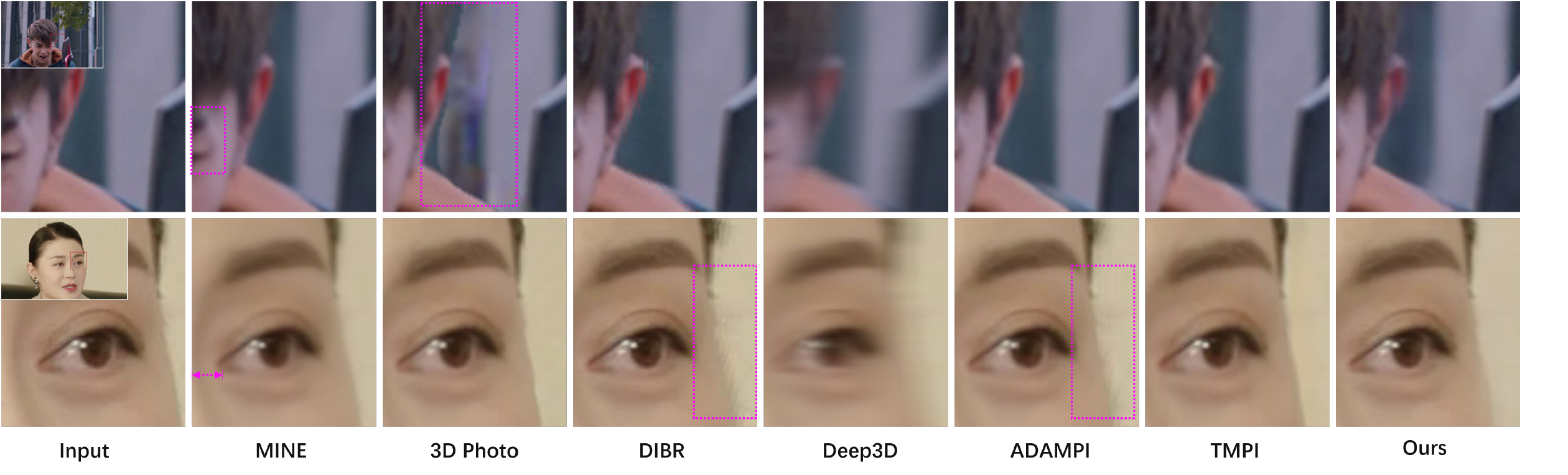}
  \caption{The planar-to-stereo conversion results of different methods for 2K planar videos.}
  \vspace{-10pt}
  \label{fig:7}
\end{figure*}
\subsubsection{3D Conversion Results of 2K Planar Videos}
The 3D video conversion at higher resolutions presents greater challenges due to the larger receptive field requirement and wider occlusion-exposure regions. Fig.\ref{fig:7} illustrates the results on the Youku2K test set. Firstly, the 3D-Photo, DIBR, and Deep3D methods still suffer from false textures, line artifacts or blurs in occlusion regions. Secondly, MINE produces smaller disparities than other methods, and ADAMPI also causes slight artifacts around the edges. Thirdly, although these methods are all trained on low-resolution images, the proposed method and TMPI do not exhibit noticeable flaws when tested on 2K images.

\begin{table}[!t]
\small
\setlength\tabcolsep{2pt}
\centering
\begin{tabular}{lcccccc}
\toprule
Method   & Extra       & MUSIQ↑                  & HIQA↑        & NIQE↓         & MOS↑    & Times(ms)   \\
\midrule
ADAMPI   & DPT         & 4.64                    & 0.367            & 5.51          & 3.40     & 422         \\
MINE     & -           & 4.66                    & 0.363            & 6.28          & 3.11    & 428          \\
3D Photo & DPT         & 4.66                    & 0.382            & 5.46          & 3.25    & 262147      \\
Deep3D   & -           & 4.45                    & 0.282            & 7.42          & 3.34    & 54          \\
DIBR     & DPT         & -                       & -                & -             & 3.21    & 11187       \\
TMPI     & DPT         & \textbf{4.67}           & 0.372            & 5.48          &\textbf{3.47}    & 721         \\
Ours     & -           & \textbf{4.67}           & \textbf{0.383 }            & \textbf{5.25} & 3.42    & \textbf{18} \\
\bottomrule
\end{tabular}
\caption{Blind image quality assessment scores and runtime of different methods on Youku2K test set.}
\vspace{-5pt}
\label{table:3}
\end{table}

\begin{table}[!t]
\small
\centering
\begin{tabular}{lcccccc}
\toprule
              & SSIM↑            & PSNR↑              & LPIPS↓ \\
\midrule
DEFF w/o detail    & 0.911             & 32.513            & 0.054  \\
DEFF w/o depth semantic  & 0.894             & 31.431            & 0.094   \\
w/o DEFF → concat      & 0.911             & 32.594            & 0.053   \\
Ours      & \textbf{0.913}        & \textbf{33.037}   & \textbf{0.037}  \\
\bottomrule
\end{tabular}
\caption{Network details ablation on 3D movie test set.}
\vspace{-3pt}
\label{table:4}
\end{table}

\begin{table}[!t]
\small
\centering
\setlength\tabcolsep{2pt}
\begin{tabular}{lcccccc}
\toprule
                     & MUSIQ↑        & HIQA↑        & NIQE↓  & Times(ms)\\
\midrule
Bilinear interpolation    & 4.42             & 0.264            & 10.81   & \textbf{16}\\
Full resolution  & \textbf{4.68 }       & \textbf{0.390}            & \textbf{5.15}  & 155 \\
Full resotion + 64 planes      & 4.67             & 0.375            & 5.80    &241\\
Ours      & 4.67        & 0.383   & 5.25   & 18\\
\bottomrule
\end{tabular}
\caption{Ablation study of simplified MPI rendering on Youku2k test set}
\vspace{-12pt}
\label{table:5}
\end{table}

Since GTs are unavailable in the 2K planar video test set, some blind image quality assessment (BIQA) results are listed in Table~\ref{table:3}, including MUSIQ \cite{ke2021musiq}, HIQA \cite{su2020blindly} and NIQE \cite{mittal2012making}. While these BIQA metrics partially reflect image quality subjectivity, they were not specifically designed for 3D conversion tasks. Therefore, these scores serve only as a one-sided reference. Additionally, we excluded the test results of DIBR due to complications arising from holes when evaluated with no-reference metrics. Notably, our proposed method achieves quality scores comparable to other larger models. To further compare the subjective quality, Table~\ref{table:3} also lists the mean opinion scores (MOS) of different methods. To obtain MOS values, 15 observers were invited to score the anonymous results in random order. The MOS scale ranges from 1 (worst) to 5 (best). These 3D results are displayed on a glasses-free 3D screen in side-by-side (SBS) format. The proposed method and TMPI obtain similar scores that are higher than other methods. Finally, the average runtime of different methods at 2K resolution is also listed in Table~\ref{table:3}, which shows the proposed method can achieve the fastest real-time inference speed. 

\subsubsection{Ablation Studies}
Ablation studies are performed on the 3D movie test set to assess the effectiveness of the designed structures. The results of the ablation tests are shown in Table~\ref{table:4}.
When the generation of depth features does not incorporate information from both two branches, particularly the depth semantic features, a significant reduction in these metrics is observed.  Additionally, replacing the DEFF module with a simple feature concatenation operation leads to a similar decrease in these metrics.
On the other hand, evaluations are carried out on the Youku2K video set to examine the simplified MPI rendering with low resolution. As shown in Table~\ref{table:5}, when processing high-resolution images, our method demonstrates superior metrics compared to the naive approach of enlarging predicted frames using bilinear interpolation. Notably, compared to running on full-resolution images, the proposed strategy significantly reduces runtime without causing a noticeable decrease in image quality scores. Moreover, increasing the number of planes to 64 lead to more difficult optimization and results in a performance decline, which can be attributed to the simplistic structure of the proposed network.

\section{Conclusion}
This paper proposed a lightweight stereoscopic conversion network based on MPI, which contains a detail branch, a depth semantic branch, and a simplified MPI rendering module. Instead of using extra depth maps, the proposed method designs a lightweight branch to calculate depth-aware features. An additional large-scale auxiliary branch is introduced to optimize the depth semantic branch in a coarse-to-fine manner, which is only used in the training phase. Experimental results indicate that the proposed approach achieves a subjective quality that is comparable to state-of-the-art methods while utilizing significantly fewer parameters and demonstrating much faster inference speed.

\bibliography{aaai25}
\end{document}